\documentclass{article}

\usepackage{amsmath,amsfonts,amssymb,times,graphicx,natbib,algorithm,algorithmic,hyperref,cleveref}

\usepackage[accepted]{hill2019}

\icmltitlerunning{Human Interaction and Interpretability Paper}

\begin{document}

\twocolumn[
\icmltitle{An Extensible Interactive Interface for Agent Design}


\icmlsetsymbol{equal}{*}

\begin{icmlauthorlist}
\icmlauthor{Matthew Rahtz}{z}
\icmlauthor{James Fang}{b}
\icmlauthor{Anca D. Dragan}{b}
\icmlauthor{Dylan Hadfield-Menell}{b}
\end{icmlauthorlist}

\icmlaffiliation{z}{Institute of Neuroinformatics, University of Z{\"u}rich and ETH Z{\"u}rich}
\icmlaffiliation{b}{Department of Electrical Engineering and Computer Sciences, University of California, Berkeley}

\icmlcorrespondingauthor{Matthew Rahtz}{mrahtz@ethz.ch}

\icmlkeywords{interpretability, transparency}

\vskip 0.3in
]


\printAffiliationsAndNotice{}

\begin{abstract}
In artificial intelligence, we often specify tasks through a reward function. While this works well in some settings, many tasks are hard to specify this way. In deep reinforcement learning, for example, directly specifying a reward as a function of a high-dimensional observation is challenging. Instead, we present an interface for specifying tasks interactively using demonstrations. Our approach defines a set of increasingly complex policies. The interface allows the user to switch between these policies at fixed intervals to generate demonstrations of novel, more complex, tasks. We train new policies based on these demonstrations and repeat the process. We present a case study of our approach in the Lunar Lander domain, and show that this simple approach can quickly learn a successful landing policy and outperforms an existing comparison-based deep RL method.
\end{abstract}

\begin{figure*}[t!]
\centering
\includegraphics[width=0.9\textwidth]{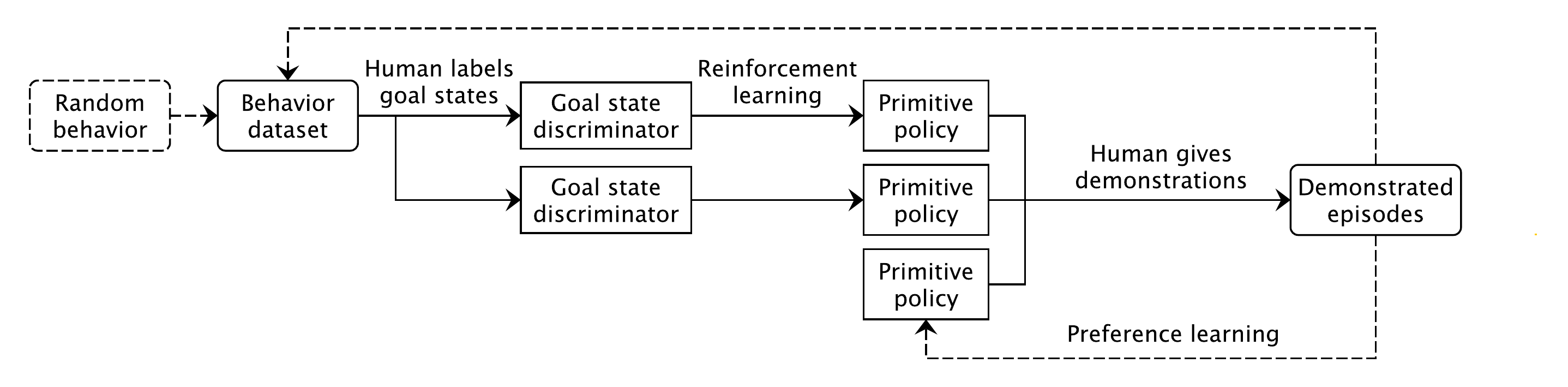}
\caption{Schematic overview of training process. Starting from behavior generated through random exploration, the user picks out a set of interesting goal states (e.g.\ robot left or right of starting position) and trains discriminators to recognize those states. Using discriminator output as a reward signal, policies are trained to reach those states consistently, creating a set of behavioral primitives (e.g.\ `move left' and `move right'). Using these primitives, the user demonstrates desired behavior. These demonstrations both add to the pool of behavior from which new goal states can be defined, and can be used to train new policies using preference learning. This process repeats until the user has trained a policy capable of performing the full task.}\label{fig:process}
\end{figure*}

\section{Introduction}

In AI we identify the correct behavior for robot systems in several ways. A popular way is through reward functions. However, reward functions make a lot of assumptions about the design setting; for example, they are only readily applicable in problems where the state space is defined in human-understandable features, or where goal states can be easily expressed in a general-purpose programming language.

Approaches based on learned reward functions allow users to specify tasks in alternative, more flexible ways. For example, in \citet{drlhp} and \citet{dorsa} the target behavior is specified through comparisons between states. Over time the reward function learns a state ranking that agrees with (and hopefully generalizes) these comparisons and is used to train a policy or optimize a trajectory. However, these methods often exhibit poor sample complexity on complex tasks, and this has led to approaches that leverage e.g. an externally generated set of expert demonstrations~\cite{ibarz2018reward}.

In this work, we draw inspiration from hierarchical planning systems where plans consist of a sequence of high-level actions, each of which runs a simpler, primitive, policy. Our approach combines two ideas: 1) we can use a set of primitive policies to efficiently define policies that perform complex tasks; and 2) in subsequent training rounds, we can use these complex policies as primitives themselves. Starting from an initial state, we sample a number of rollouts of fixed length from our primitive policies. We show these rollouts to a human designer who selects the best one for the target task in an interface akin to those used in~\citet{drlhp} and \citet{dorsa}. The process then repeats from the final state in the selected rollout until the designer indicates the end of an episode.

From this data, we can train new policies in two ways. First, from the implicit comparisons between the best rollout and the other rollouts, we learn a reward function using preference learning. This reward function can then be used to train a policy using reinforcement learning. Second, we can train goal classifiers based on states reached in the demonstrations, and train a policy using reinforcement learning to maximize goal classification probability. This provides the user with a way to specify behaviors quickly when arriving at a goal state is easy through a sub-optimal demonstration.

We provide a demonstration of this approach in the Lunar Lander domain. We train policies in three stages. In the first stage, we use random behavior to learn two policies: stabilize and drop. Stabilize is trained to reach goal states where the lander is level and (close to) stationary. Drop is trained to turn off the engines for final landing. In the second stage, we use a combination of random behavior and the stabilize policy to train policies that move stably to the left or stably to the right. In the final stage, we use all four policies to train a policies that successfully lands without crashing. Our final solution is able to land successfully in over 90\% of episodes.

\section{Related work}

This work continues a recent trend of agent specification using interpretable techniques.

One example is \emph{preference learning}, where specification is based on comparisons between examples of behavior. However, in preference learning, the user has little control over exploration---that is, what kind of examples they are comparing, and therefore what kind of information they are supplying. Existing approaches for generating examples include selection based on reward uncertainty~\cite{drlhp} and active synthesis of maximally-informative comparisons~\cite{dorsa}. In contrast, our approach allows the user to influence exploration directly through demonstration.

Another example of this trend is \emph{natural language instruction}. However, natural language must be grounded in the environment. This can be difficult; possibilities include demonstration of instructions~\cite{gml}, examples of goal states corresponding to instructions~\cite{agile}, and rewards manually conditioned on instructions~\cite{grounded}. We approach the problem from the opposite direction: instead of taking an existing vocabulary and grounding it in the environment, we give the user the means to define a vocabulary of behaviors for themselves.

A final example is specification through \emph{goal states}~\cite{uvfa,her,nair2018visual,vice,viceraq}. Goal states provide an accessible alternative to reward engineering in cases where examples of the final goal can be generated easily. We extend this line of work to cases where the final goal is initially difficult or impossible to reach (such as the final state of a tricky game), and where therefore an iterative approach is required, incorporating a higher degree of human feedback than goal states alone.

\section{Method}

Training is based on demonstrations using a set of behaviors defined by the user. These behaviors, which we term \emph{behavioral primitives}, are encoded as policies. The training process is iterative: experience generated by one set of demonstrations is used to define primitives for the next set of demonstrations, continuing until a primitive that can perform the full task. The user defines the first set of primitives by identifying interesting goal states in experience generated by a random policy then training policies to reach each of those states. Thereafter the user defines primitives either based either on goal states in experience generated by previous demonstrations, or based directly on behaviors demonstrated. The training system therefore consists of three components:

\begin{itemize}
    \item An interface for defining goal states based on experience generated during training so far.
    \item Apparatus for training behavioral primitives.
    \item An interface for giving demonstrations using those behavioral primitives.
\end{itemize}

Each of these components is described in detail below.

\subsection{Defining goal states}

One way to define a behavioral primitive is by training a goal-conditioned policy based on a goal defined by the user. Early work in goal-conditioned RL focused on goals parameterized using specific environment states~\cite{uvfa,her,nair2018visual}. In contrast, we continue recent work~\cite{vice,viceraq} employing goal states based on abstract concepts. Instead of referring to, say, a particular position in the room, one of our goal states might be `near a human', yielding a behavioral primitive that moves to a human. This enlarges the set of possible behaviors that can be encoded as goal-conditioned policies. The user defines these abstract goal states by example. These examples are used to train a binary classifier (which we refer to as a \emph{discriminator}) to recognize whether the agent is in the defined state.

The user begins by browsing videos of previous episodes generated during the training process. Initially, these episodes are generated by a random policy, but later in training these include episodes generated by demonstrations. Once the user has identified an `interesting' state (a behavior she believes will be useful for later demonstrations), she labels video frames as positive or negative examples of that state. Frames are mapped to environment observations (which may be lower-dimensional than the video frames) by the system, creating a set of positive and negative examples of observations. These examples are then used to train a binary classifier, a \emph{discriminator}, implemented using a neural network, to recognise the abstract state. In our experiments below, roughly 400 examples are required to train a robust discriminator for each state.

\subsection{Behavioral primitives from goal states}

\begin{figure}[!t]
\centering
\includegraphics[width=0.5\textwidth]{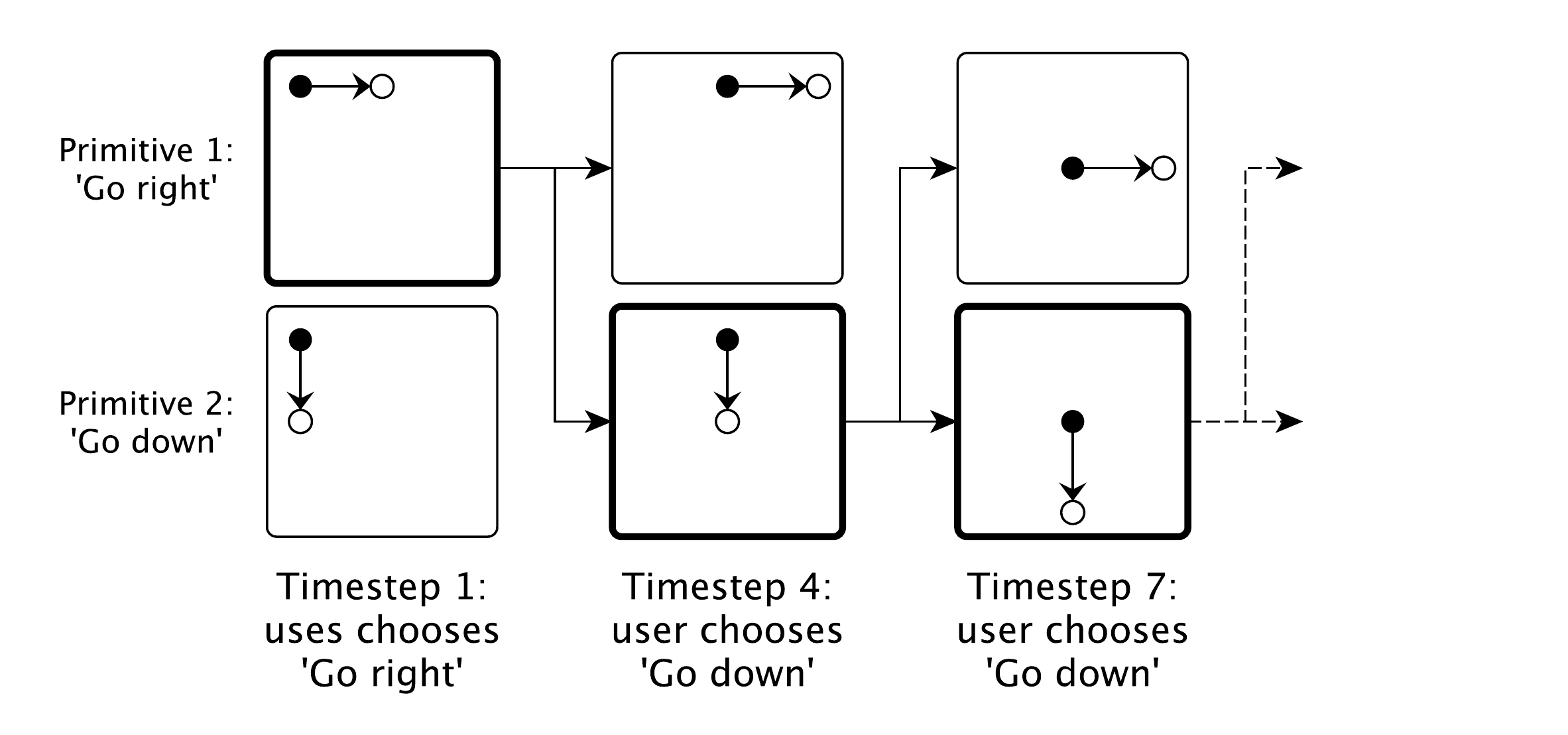}
\caption{Demonstrations interface. Demonstrations are given one action at a time. Each action corresponds to some temporally-extended behavior generated by running the corresponding primitive policy for a fixed number of timesteps (here, three timesteps). Because the results of each primitive may be unpredictable, the user chooses between actions based on the simulated rollouts that would result (here, the filled circle moving to the position indicated by the unfilled circle). Once the user chooses an action, the final state from the corresponding rollout is used as the first state of the next set of rollouts.}\label{fig:interface}
\end{figure}

To define primitive policies based on user-defined goal states, we use discriminator probability output as a reward signal then train using an off-the-shelf reinforcement learning algorithm, Proximal Policy Optimization~\cite{ppo} from OpenAI Baselines~\cite{baselines}. To maximise reward, the resulting policy must activate the discriminator as strongly as possible - in other words, it must move to and stay in the goal state.

\subsection{Demonstrations using behavioral primitives}

The user gives demonstrations by using primitive policies as temporally-extended actions. Based on the current environment state, the user chooses a policy; the policy is run for a fixed number of timesteps; based on the new state, the user chooses a new policy; and so on. Primitive policies are thus similar to \emph{options} in the options framework~\cite{options}, with a termination condition based on the number of steps.

When the effect of each primitive is predictable, the choice of primitive at each step is straightforward. In general, however, we assume primitives will be somewhat \textit{un}predictable. This is partly because we do not expect trained primitives to be perfect (e.g. movement may be erratic). However, the environment itself may also be unpredictable. In the Lunar Lander game described below, for example, it is difficult to predict how the spacecraft will move in low gravity. To enable an informed choice, at each demonstration step we show the user not only the current environment state but also a video of the rollout that would result from running each policy from that state. Essentially, the user chooses by examining the short-term futures that would result from each action.

This interface is illustrated in Figure~\ref{fig:interface}.

\subsection{Behavioral primitives from demonstrations}

In addition to defining primitives from goal states, we also support defining behavioral primitives directly from behavior demonstrated by the user. Our early experiments used a simple imitation learning technique, behavioral cloning~\cite{bc}. However, the resulting policies would often perform significantly worse than the demonstrations themselves. Instead, we note that our demonstrations offer an additional source of information. Each demonstrated action yields not only information about optimal behavior (the rollout from the selected primitive) but also \emph{comparisons} to sub-optimal behaviors from the same state (the rollouts from the other primitives). These comparisons can be used to train a policy using \emph{preference learning} techniques.

In particular, we implement preference learning based on~\citet{drlhp}. This involves training a neural network to predict the result of each pairwise comparison (between the chosen rollout and one of the other rollouts, predicting which was the chosen rollout). The prediction is made using a latent reward value calculated for each frame in each rollout. This predicted reward function is then used to train a policy using a reinforcement learning algorithm. Again, we use Proximal Policy Optimization~\cite{ppo}.

(We also investigating combining behavioral cloning with preference learning, but we found this to result in worse performance than preference learning alone. Future work will investigate other ways of making use of both types of information.)

\subsection{Iterated training}

For simple tasks, it may be possible to give demonstrations of the full task using only the first set of primitives defined, and from those demonstrations train a final behavioral primitive that successfully performs the task. In general, however, we assume that initial primitives will only enable demonstration of some part of the task.

The user has a number of options for defining new primitives based on previous primitives.

\textbf{Demonstration of new behavior}. The user might directly demonstrate a behavior she intends to use in later demonstrations, distilling those demonstrations into a new primitive policy as described above. For example, using a set of basic quadcopter movement primitives the user might demonstrate a loop-the-loop behavior and use this as a new primitive.

\textbf{Goal states from demonstrations}. Demonstrations generate experience exploring parts of the state space not covered by the initial set of random behavior. From this experience new goal states can be defined from which new primitives can be trained. Returning to the quadcopter example, the user could demonstrate moving the quadcopter over a charging platform, and using that goal state, train a policy to move to the charging platform. Practically, this involves saving demonstrated episodes for browsing and labelling with the same interface as the initial set of goal states.

\textbf{Curriculum learning with goal states}. In some cases it may not be necessary to \emph{demonstrate} exploration of new parts of the state space; interesting new states may be reached by simply running one of the existing primitive policies in the environment (assuming that the policy is stochastic and therefore does some exploration of its own). Consider training a robot to navigate a maze. Random exploration from the initial state may only wander around in the first part of the maze, so that the best goal state initially possible to define is only a short distance along the correct path. But by exploring randomly in the vicinity of this first state, it is easier to wander into a deeper part of the maze in which a second goal state may be defined, and so on. Essentially, we can train using a \emph{curriculum} of gradually more advanced goals. Running a single policy in the environment can be seen as a special type of demonstration where only one action is available.

This full training process is shown in \Cref{fig:process}.

\section{Case study: Lunar Lander}

\begin{figure}[!t]
\centering
\includegraphics[width=0.45\textwidth]{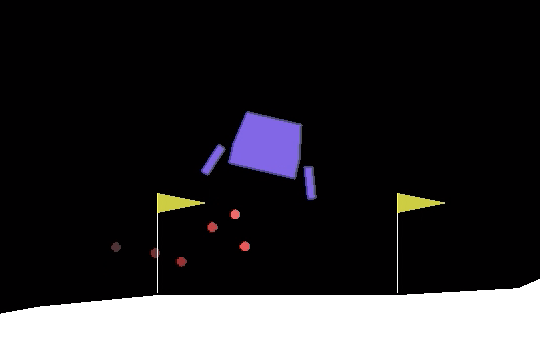}
\caption{Lunar Lander gameplay. The user must guide the descent of the spacecraft, landing in the area designated by the two flags.}
\label{fig:lunarlander}
\end{figure}

As a concrete example, we use this system to train an agent to play a video game. Lunar Lander is a simple game included with OpenAI Gym~\cite{gym} in which a user must control a 2D spacecraft, landing gently in a designated landing zone on the lunar surface (\Cref{fig:lunarlander}). With a (discrete) action space of `rotate spacecraft left', `rotate spacecraft right' and `fire thrusters at the bottom of the spacecraft', the spacecraft is very hard to control; it is challenging to train a robust policy through simple imitation learning because it is difficult to give good demonstrations. Previous work with Lunar Lander has, for example, attempted to make control easier by assisting the user in the original action space~\cite{sharedautonomy}. In contrast, we enable the user to define a \emph{new} control space which is easier to use in the first place.

\subsection{Goal states and behavioral primitives}

\begin{figure}[!ht]
\centering
\includegraphics[width=0.5\textwidth]{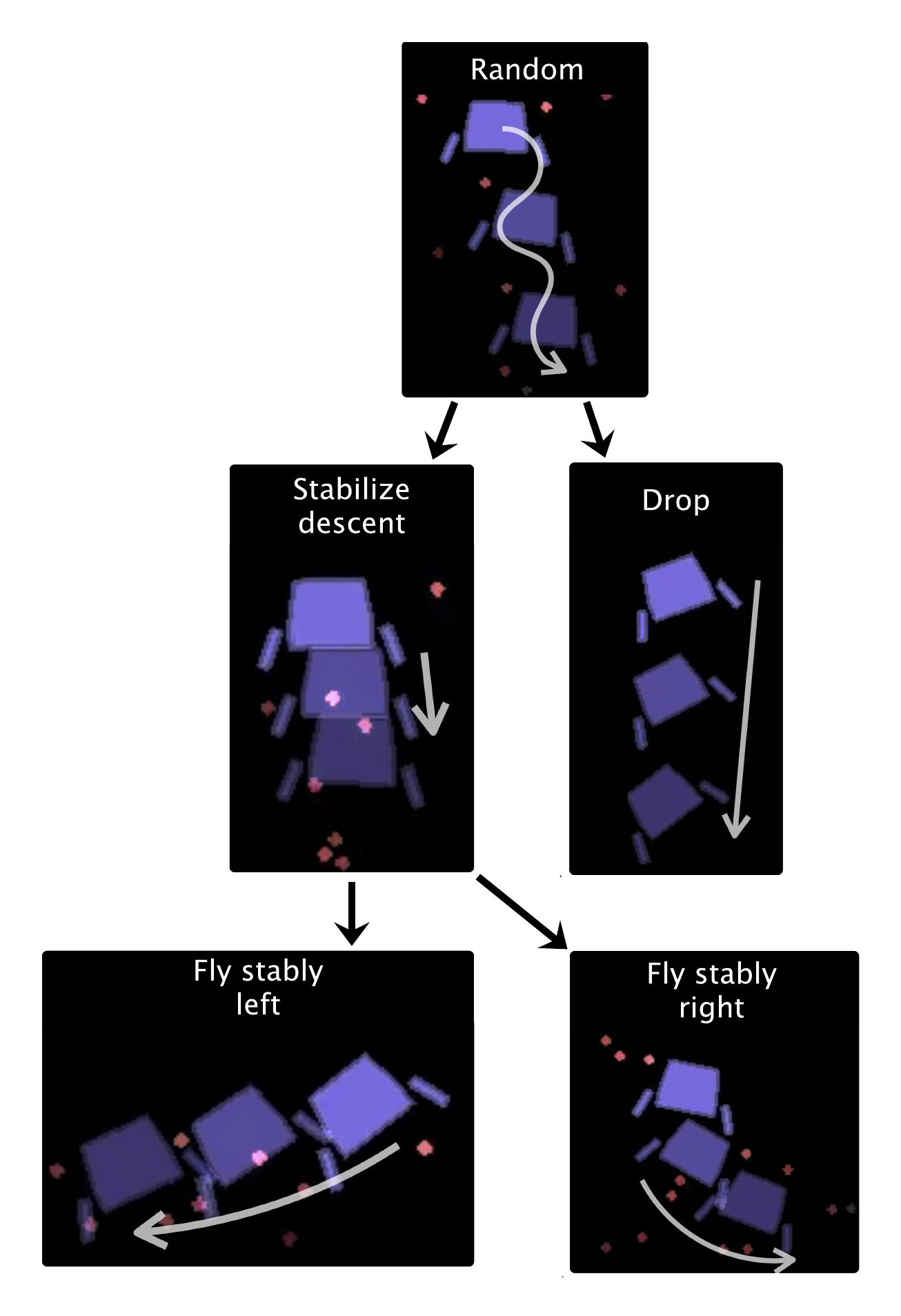}
\caption{Primitives defined for Lunar Lander. Starting from random behavior, we use goal states to define `stabilize descent' and `drop with engines off' primitives. We then define further goal states in experience generated by the `stabilize descent' primitive to define `fly stably left' and `fly stably right' primitives.}
\label{fig:primitives}
\end{figure}

\textbf{Iteration 1}: starting from random behavior, we define a \textbf{stabilize descent} goal state and primitive. Each episode begins with the spacecraft falling towards the surface at a random speed and angle; this primitive slows down the spacecraft and returns its angle to neutral. We train this primitive based on examples of the spacecraft being level and having a low velocity (angle and velocity are both included in the observation space).

\textbf{Iteration 2}: from experience generated by running the `stabilize descent' policy, we define \textbf{stably fly left} and \textbf{stably fly right} primitives. One major difficulty when demonstrating with original controls is the need to rotate in order to move left or right. It is easy to rotate too much and become unstable. These primitives move the spacecraft slowly left or right without allowing the angle to deviate too much from neutral. We also train these primitives using goal states. We generate experience using the `stabilize descent' primitive, which is not perfect and sometimes drifts slightly in one direction. We collect examples of this drifting and use those examples to train goal state discriminators and corresponding policies.

\textbf{Iteration 3}: we define a \textbf{drop} primitive, completely shutting off the spacecraft's engines so that when the craft is sufficiently close to the lunar surface we can actually land. This primitive is defined based on instances from the initial set of random behavior in which the engine is not firing.

An illustration of these primitives is shown in \Cref{fig:primitives}.

\textbf{Iteration 4}: finally, using these four primitives---`stabilize descent', `fly stably left', `fly stably right' and `drop'---we train a policy to actually play the game by demonstrating successful landings and training a policy from those demonstrations.

Defining this set of four demonstration primitives takes roughly one hour. Though the primitives are not perfect, they are sufficient to land the spacecraft in the landing zone in 80\% of demonstrations.

\subsection{Results}

\begin{figure}[t!]
\centering
\includegraphics[width=0.5\textwidth]{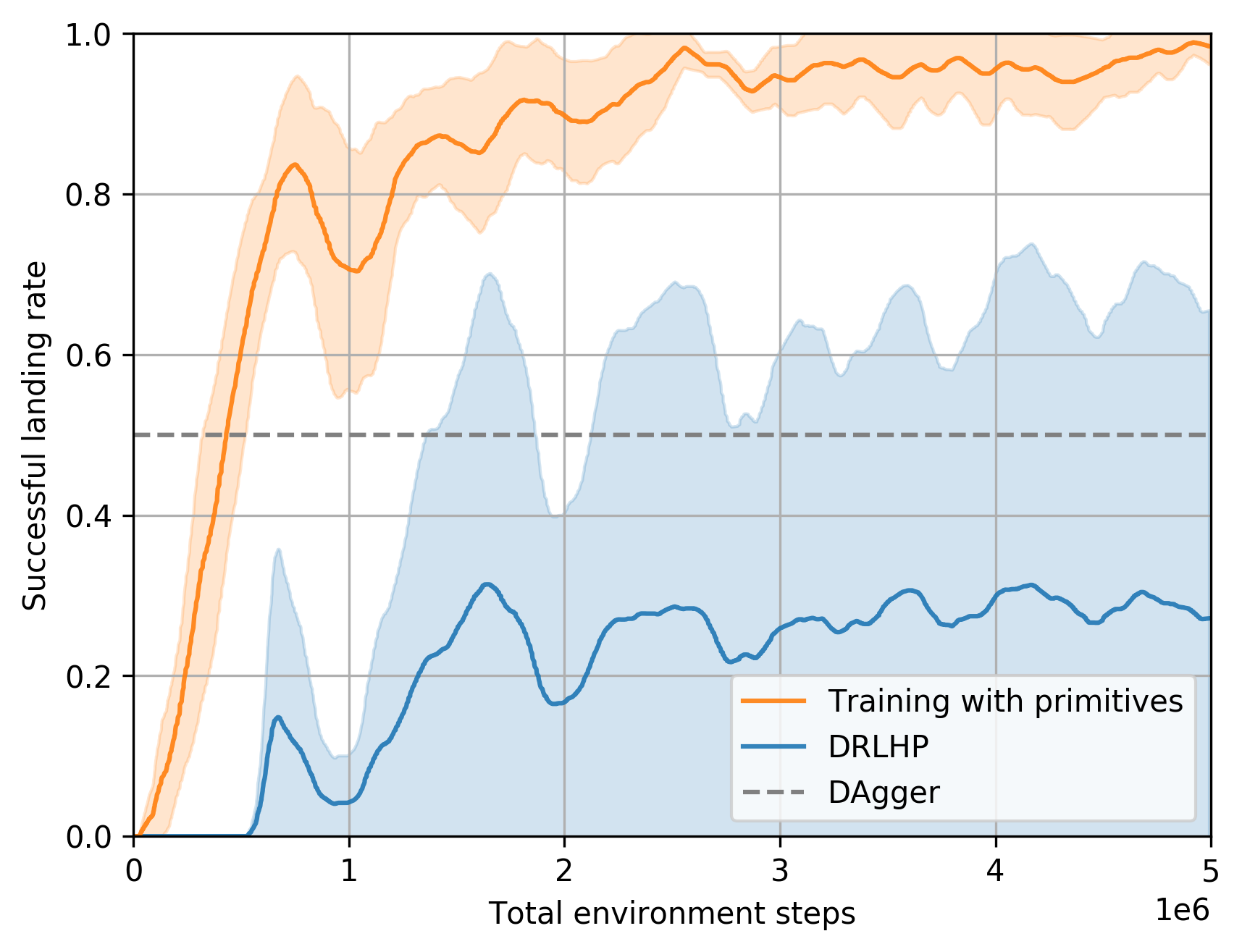}
\caption{Lunar Lander training results---success rate against number of RL steps in the environment. Using our approach, we are able to train a policy which lands successfully in almost all episodes. A comparable preference learning baseline, Deep Reinforcement Learning from Human Preferences~\cite{drlhp}, succeeds in only one out of three runs. A baseline using imitation learning instead of reinforcement learning, DAgger~\cite{dagger}, is more robust but does not achieve high performance. (Successful landing rate is calculated using a moving window of ten episodes. Shaded regions indicate one standard deviation across three runs, each with a different random seed. Full training curve for DAgger not shown due to dissimilarity in training method.)}
\label{fig:results}
\end{figure}

Starting from the four demonstration primitives described above, we perform three training runs, each with a different random initialization. In all runs, we found that only eight demonstrated episodes are required to train a successful policy. This requires roughly 15 minutes of human interaction time, followed by 90 minutes of training time (reinforcement learning using the reward function inferred from demonstrations). The resulting policy is capable of landing successfully in over 90\% of episodes (see \Cref{fig:results}).

We compare these results against two baselines, matching 15 minutes of human interaction time in each case. First, we compare to preference learning from randomly-selected examples of behavior generated while training, as in~\citet{drlhp}. Here, training is significantly less robust; we could only train a successful policy in one out of three runs. In the other two runs, the policy did not produce the kind of examples that would enable the user to give informative preferences, resulting in the policy only learning to hover in mid-air rather than to land. Second, we compare to simple imitation learning using Dataset Aggregration (DAgger)~\cite{dagger}. Here the trained policy does learn to land, but often does so by crash-landing, and often misses the target area.

\section{Conclusions and discussion}

We have presented a proposal for an interactive training interface that combines several learning methods to enable a non-technical user to build a policy from scratch. We use this interface to build a policy that plys the Lunar Lander game, and find that our method outperforms two comparable baseline methods in robustness and policy performance.

\subsection{Future work}

\textbf{Incorporating imitation learning}. We were surprised to find that combining behavioral cloning with preference learning resulted in worse performance than preference learning alone. We would like to explore alternative methods of combining the two---for example, using behavioral cloning to pre-train, rather than as part of a combined loss.

\textbf{Variable-timescale primitives}. In this work, all primitives ran for the same number of timesteps. However, part of the power of an iterated training process is that demonstrations can take place at increasingly abstract levels as training progresses. To enable this, primitive policies would need to run for more steps. There should be a clear criterion for how many steps are necessary for each policy. One way to achieve this might be through flexible termination conditions, as used in the options framework~\cite{options}.

\subsection{Acknowledgements}

We thank Rohin Shah and Adam Gleave for helpful comments and suggestions on an earlier version of this manuscript. This work was supported by the Center for Human-Compatible AI.

\bibliography{main}

\begin{thebibliography}{18}
\providecommand{\natexlab}[1]{#1}
\providecommand{\url}[1]{\texttt{#1}}
\expandafter\ifx\csname urlstyle\endcsname\relax
  \providecommand{\doi}[1]{doi: #1}\else
  \providecommand{\doi}{doi: \begingroup \urlstyle{rm}\Url}\fi

\bibitem[Andrychowicz et~al.(2017)Andrychowicz, Wolski, Ray, Schneider, Fong,
  Welinder, McGrew, Tobin, Abbeel, and Zaremba]{her}
Andrychowicz, M., Wolski, F., Ray, A., Schneider, J., Fong, R., Welinder, P.,
  McGrew, B., Tobin, J., Abbeel, O.~P., and Zaremba, W.
\newblock {Hindsight Experience Replay}.
\newblock In \emph{Advances in Neural Information Processing Systems}, pp.\
  5048--5058, 2017.

\bibitem[Bahdanau et~al.(2018)Bahdanau, Hill, Leike, Hughes, Kohli, and
  Grefenstette]{agile}
Bahdanau, D., Hill, F., Leike, J., Hughes, E., Kohli, P., and Grefenstette, E.
\newblock {Learning to Follow Language Instructions with Adversarial Reward
  Induction}.
\newblock \emph{arXiv:1806.01946}, 2018.

\bibitem[Brockman et~al.(2016)Brockman, Cheung, Pettersson, Schneider,
  Schulman, Tang, and Zaremba]{gym}
Brockman, G., Cheung, V., Pettersson, L., Schneider, J., Schulman, J., Tang,
  J., and Zaremba, W.
\newblock {OpenAI Gym}.
\newblock \emph{arXiv:1606.01540}, 2016.

\bibitem[Christiano et~al.(2017)Christiano, Leike, Brown, Martic, Legg, and
  Amodei]{drlhp}
Christiano, P., Leike, J., Brown, T., Martic, M., Legg, S., and Amodei, D.
\newblock {Deep Reinforcement Learning from Human Preferences}.
\newblock In \emph{Advances in Neural Information Processing Systems}, pp.\
  4299--4307, 2017.

\bibitem[Co-Reyes et~al.(2018)Co-Reyes, Gupta, Sanjeev, Altieri, DeNero,
  Abbeel, and Levine]{gml}
Co-Reyes, J., Gupta, A., Sanjeev, S., Altieri, N., DeNero, J., Abbeel, P., and
  Levine, S.
\newblock {Guiding Policies with Language via Meta-Learning}.
\newblock \emph{arXiv:1811.07882}, 2018.

\bibitem[Dhariwal et~al.(2017)Dhariwal, Hesse, Klimov, Nichol, Plappert,
  Radford, Schulman, Sidor, Wu, and Zhokhov]{baselines}
Dhariwal, P., Hesse, C., Klimov, O., Nichol, A., Plappert, M., Radford, A.,
  Schulman, J., Sidor, S., Wu, Y., and Zhokhov, P.
\newblock {OpenAI Baselines}.
\newblock \url{https://github.com/openai/baselines}, 2017.

\bibitem[Fu et~al.(2018)Fu, Singh, Ghosh, Yang, and Levine]{vice}
Fu, J., Singh, A., Ghosh, D., Yang, L., and Levine, S.
\newblock {Variational Inverse Control with Events: A General Framework for
  Data-Driven Reward Definition}.
\newblock In \emph{Advances in Neural Information Processing Systems}, pp.\
  8538--8547, 2018.

\bibitem[Hermann et~al.(2017)Hermann, Hill, Green, Wang, Faulkner, Soyer,
  Szepesvari, Czarnecki, Jaderberg, Teplyashin, et~al.]{grounded}
Hermann, K.~M., Hill, F., Green, S., Wang, F., Faulkner, R., Soyer, H.,
  Szepesvari, D., Czarnecki, W.~M., Jaderberg, M., Teplyashin, D., et~al.
\newblock {Grounded Language Learning in a Simulated 3D World}.
\newblock \emph{arXiv:1706.06551}, 2017.

\bibitem[Ibarz et~al.(2018)Ibarz, Leike, Pohlen, Irving, Legg, and
  Amodei]{ibarz2018reward}
Ibarz, B., Leike, J., Pohlen, T., Irving, G., Legg, S., and Amodei, D.
\newblock {Reward learning from human preferences and demonstrations in Atari}.
\newblock In \emph{Advances in Neural Information Processing Systems}, pp.\
  8022--8034, 2018.

\bibitem[Nair et~al.(2018)Nair, Pong, Dalal, Bahl, Lin, and
  Levine]{nair2018visual}
Nair, A., Pong, V., Dalal, M., Bahl, S., Lin, S., and Levine, S.
\newblock Visual reinforcement learning with imagined goals.
\newblock In \emph{Advances in Neural Information Processing Systems}, pp.\
  9208--9219, 2018.

\bibitem[Pomerleau(1991)]{bc}
Pomerleau, D.
\newblock {Efficient Training of Artificial Neural Networks for Autonomous
  Navigation}.
\newblock \emph{Neural Computation}, 3\penalty0 (1):\penalty0 88--97, 1991.

\bibitem[Reddy et~al.(2018)Reddy, Dragan, and Levine]{sharedautonomy}
Reddy, S., Dragan, A.~D., and Levine, S.
\newblock {Shared Autonomy via Deep Reinforcement Learning}.
\newblock \emph{arXiv:1802.01744}, 2018.

\bibitem[Ross et~al.(2011)Ross, Gordon, and Bagnell]{dagger}
Ross, S., Gordon, G., and Bagnell, D.
\newblock {A Reduction of Imitation Learning and Structured Prediction to
  No-Regret Online Learning}.
\newblock In \emph{{Proceedings of the Fourteenth International Conference on
  Artificial Intelligence and Statistics}}, pp.\  627--635, 2011.

\bibitem[Sadigh et~al.(2017)Sadigh, Dragan, Sastry, and Seshia]{dorsa}
Sadigh, D., Dragan, A., Sastry, S., and Seshia, S.
\newblock {Active Preference-Based Learning of Reward Functions}.
\newblock In \emph{Robotics: Science and Systems}, 2017.

\bibitem[Schaul et~al.(2015)Schaul, Horgan, Gregor, and Silver]{uvfa}
Schaul, T., Horgan, D., Gregor, K., and Silver, D.
\newblock {Universal Value Function Approximators}.
\newblock In \emph{{International Conference on Machine Learning}}, pp.\
  1312--1320, 2015.

\bibitem[Schulman et~al.(2017)Schulman, Wolski, Dhariwal, Radford, and
  Klimov]{ppo}
Schulman, J., Wolski, F., Dhariwal, P., Radford, A., and Klimov, O.
\newblock {Proximal Policy Optimization Algorithms}.
\newblock \emph{arXiv:1707.06347}, 2017.

\bibitem[Singh et~al.(2019)Singh, Yang, Hartikainen, Finn, and Levine]{viceraq}
Singh, A., Yang, L., Hartikainen, K., Finn, C., and Levine, S.
\newblock {End-to-End Robotic Reinforcement Learning without Reward
  Engineering}.
\newblock \emph{arXiv preprint arXiv:1904.07854}, 2019.

\bibitem[Sutton et~al.(1999)Sutton, Precup, and Singh]{options}
Sutton, R., Precup, D., and Singh, S.
\newblock {Between MDPs and semi-MDPs: A framework for temporal abstraction in
  reinforcement learning}.
\newblock \emph{Artificial intelligence}, 112\penalty0 (1-2):\penalty0
  181--211, 1999.

\end{thebibliography}
\bibliographystyle{icml2019}

\end{document}